\documentclass[final]{cvpr}

\usepackage{times}
\usepackage{epsfig}
\usepackage{graphicx}
\usepackage{amsmath}
\usepackage{amssymb}
\usepackage{flexisym}
\usepackage{booktabs}
\usepackage{multirow}
\usepackage{array}
\usepackage{tabularx}
\usepackage{ragged2e}
\usepackage{multicol}
\usepackage{lineno}
\usepackage{subfigure}

\usepackage[pagebackref=true,breaklinks=true,colorlinks,bookmarks=false]{hyperref}

\begin{document}

%%%%%%%%% TITLE
\title{Exploring Self-Attention for Visual Odometry}

\author{Hamed Damirchi, Rooholla Khorrambakht, Hamid D. Taghirad, Senior Member, IEEE \\
Faculty of Electrical and Computer Engineering\\
K. N. Toosi University of Technology\\
P.O. Box 16315-1355, Tehran, lran\\
{\tt\small hdamirchi, r.khorrambakht@email.kntu.ac.ir, taghirad@kntu.ac.ir}
% For a paper whose authors are all at the same institution,
% omit the following lines up until the closing ``}''.
% Additional authors and addresses can be added with ``\and'',
% just like the second author.
% To save space, use either the email address or home page, not both
% \and
% Second Author\\
% Institution2\\
% First line of institution2 address\\
% {\tt\small secondauthor@i2.org}
}

\maketitle

%%%%%%%%% ABSTRACT
\begin{abstract}
   Visual odometry networks commonly use pretrained optical flow networks in order to derive the ego-motion between consecutive frames. The features extracted by these networks represent the motion of all the pixels between frames. However, due to the existence of dynamic objects and texture-less surfaces in the scene, the motion information for every image region might not be reliable for inferring odometry due to the ineffectiveness of dynamic objects in derivation of the incremental changes in position. Recent works in this area lack attention mechanisms in their structures to facilitate dynamic reweighing of the feature maps for extracting more refined egomotion information. In this paper, we explore the effectiveness of self-attention in visual odometry. We report qualitative and quantitative results against the SOTA methods. Furthermore, saliency-based studies alongside specially designed experiments are utilized to investigate the effect of self-attention on VO. Our experiments show that using self-attention allows for the extraction of better features while achieving a better odometry performance compared to networks that lack such structures.
   
\end{abstract}

%%%%%%%%% BODY TEXT
\section{Introduction}

\begin{figure}[t]
   \includegraphics[width=0.98\linewidth]{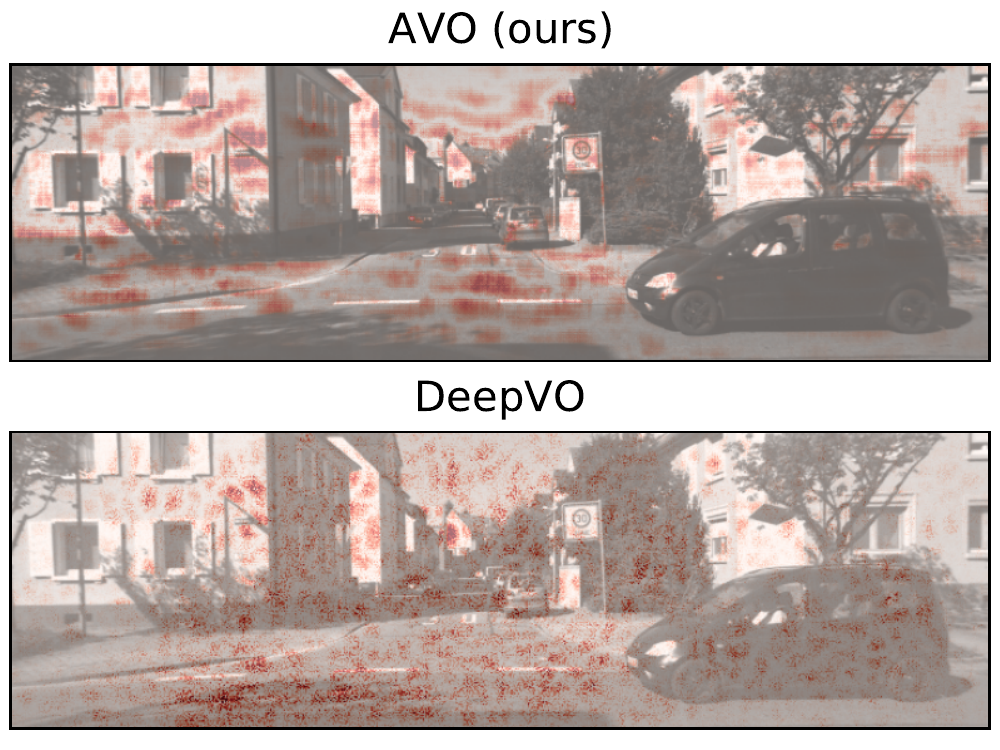}
   \caption{Saliency maps of the 7th sequence of the KITTI~\cite{geiger2012kitti} dataset. The difference between the two maps show that our method is able to reject the features corresponding to moving objects while DeepVO~\cite{wang2017deepvo} shows no reactions to the presense of such artifacts.}
   \label{fig:firstpageattr}
\end{figure}

Visual odometry (VO) refers to the task of deriving the incremental poses between consecutive frames captured by a camera. Over the past decade, VO algorithms have received attention from the robotics \cite{Forster2016}, and especially autonomous driving \cite{chen2020survey, milz2018sdvo} research communities, where vision-based subsystems have played a crucial role as localization pipelines. Classical approaches to visual odometry rely on hand-crafted methods to extract features from consecutive images. However, This process is not robust to real-world artifacts such as intense illumination changes and near texture-less surfaces. Moreover, the extracted features do not encode any semantic meanings that may provide clues about motion patterns of objects and their reliability for VO. Therefore, such classical methods fail when dynamic objects are present in the scene or when the feature extraction methods fail to extract enough distinctive descriptors for an adequate ego-motion estimation.

Deep learning has been proposed as a means to provide robust feature extractors \cite{sagawa2016robust}. Convolutional neural networks (CNN) are pervasively used to extract rich features from a single image or a stream of images in various fields such as optical flow estimation~\cite{ilg2016flownet}, and next-frame prediction networks~\cite{Villegas2017DecomposingMA}. Traditionally, CNNs utilize local convolutions in order to extract features from different regions of the input. This locality means that they are not able to globally manipulate the extracted features from one region at every stage, based on the information derived from other areas of the same image. In other words, CNNs lack structures that provide information about the content of the image on a global scale rather than a local one. Hence, in the case of VO networks, the local content at each stage may not reject the unreliable features corresponding to dynamic objects that are present in the image.

Self-attention \cite{wang2017nonlocal} has been proposed as a non-local method that allows the network to pool information from various regions of the input in order to provide a means for the manipulation of the data at a local level. Hence, self-attention layers allow each pixel to attend to all pixels regardless of their position on the feature map. Such methods have been applied to relocalization \cite{wang2020atloc}, yet, no rigorous exploration has been carried out to investigate their efficacy and structural benefits for visual odometry. In this paper, we explore the benefits of utilizing self-attention mechanisms in VO networks by focusing on issues such as feature importance and moving objects. We provide extensive visualizations to illustrate the ability of attention based VO algorithms to consistently focus on salient features from the images and reject dynamic objects as can be seen in Fig.~\ref{fig:firstpageattr}.  Self-attention based VO, while requiring minimal architectural modifications and added parameters, allows us to achieve higher accuracies in egomotion estimation compared to attention-less SOTA networks. Moreover, Our comparisons point out the lack of structure in the process of feature extraction capabilities of attention-less networks. This paper is structured as follows. We provide a brief overview of the current SOTA in VO and commonly used attention modules in Section \ref{sec:relatedwork}. We present our method in Section \ref{sec:method} and provide details of the proposed architecture. Section \ref{implementation} discusses the setting of our experiments alongside implementation details, and in Section \ref{experiments}, we present the quantitative and qualitative results alongside providing custom evaluations that further elucidate the benefit of attention in VO. Our contributions are as follows:
\begin{itemize}
    \item Provide quantitative analysis to show the boost in performance when using self-attention in VO and compare our method against SOTA, VO methods that lack attention mechanisms
    \item Visualize the temporal frame-to-frame losses, directly, for attention-enabled and attention-less methods when dynamic objects are present in the scene
    \item Provide gradient-based analyses to qualitatively analyze the effect of self-attention on the feature extraction capabilities of the network
    \item Provide temporal analyses based on the semantic segmentation maps from the input images to show the consistency of our method in attending to salient features from various object categories
\end{itemize}

%-------------------------------------------------------------------------
\section{Related Work}
\label{sec:relatedwork}
In this section, we provide an overview of VO and localization methods available in the literature and the various attention mechanisms used to improve the performance of these pipelines. Algorithmically, localization may be categorized into two broad approaches, odometry and relocalization. Odometry focuses on deriving the incremental changes in position, while relocalization is concerned with treating the global pose of inputs as a regression problem. Unlike learning-based methods that capture the motion model directly from data, classical methods utilize scene geometry and iterative non-linear optimizers to perform odometry.

In classic visual odometry methods, a common assumption is that the world around is static \cite{zhang2020vdoslam}. Furthermore, the simplifications considered in deriving the camera models in classical methods and the lack of learning to compensate for them limit the deployability of a large portion of classical methods. Such pipelines deal with moving objects in the scene through outlier detection methods and robust optimizers. As a different approach, multi-object tracking may also be utilized to remove these objects from the images~\cite{wangsiripitak2009movingobject}. In this paper, we propose an end-to-end method that is able to extract adequate features while disregarding the information from moving objects without any preprocessing or deliberate augmentations of data during training.

DeepVO \cite{wang2017deepvo} was the first learning-based visual odometry method that used a CNN-LSTM approach and achieved competitive results against classical VO methods. This work was later extended \cite{wang2018espvo} to include uncertainty alongside the pose outputs from the network. Neither of the mentioned methods modified the CNN architecture and used a network pretrained on optical flow to extract features from consecutive frames. In later sections, we show that compared to this method, better performance is achieved by modifying the visual feature extractor using attention layers. Moreover, due to the similarity of the base structure of our network to \cite{wang2017deepvo}, we are able to compare our attention-based VO network directly against this SOTA attention-less network.

\cite{wang2020atloc} proposes AtLoc, a relocalization system that uses attention mechanisms in order to increase the performance of the relocalization pipelines. This work shows that a trained attention-less network uses regions of the input that are not suitable for performing relocalization. Instead, they focus on impermanent foreground features rather than the background objects which should be the main factor when inferring the global location of the input.

\cite{fei2019beyond} proposes the adoption of external memory modules with VO pipelines and introduces a pose refinement approach based on temporal attention layers in order to rectify the pose features at different iterations of the algorithm. In our approach, we propose the utilization of self-attention mechanisms to refine the extracted motion features from the visual encoder of the network.

\cite{kuo2020davo} proposes two attention-based VO methods. the first method utilizes a segmentation network to provide semantic masks that weigh the different regions of the image with the aid of an optical flow network. The second proposed network in \cite{kuo2020davo} uses squeeze-and-excite~\cite{hu2018seblock} attention blocks. In this paper, we utilize spatial self-attention rather than channel-wise attention while no auxiliary networks are used to provide attention over the feature maps. Furthermore, we analyze the effectiveness of such attention mechanisms in different scenarios and provide visualizations and in-depth analyses to explain the necessity of spatial self-attention in the task of VO.

\begin{figure*}[t]
   \includegraphics[width=0.98\linewidth]{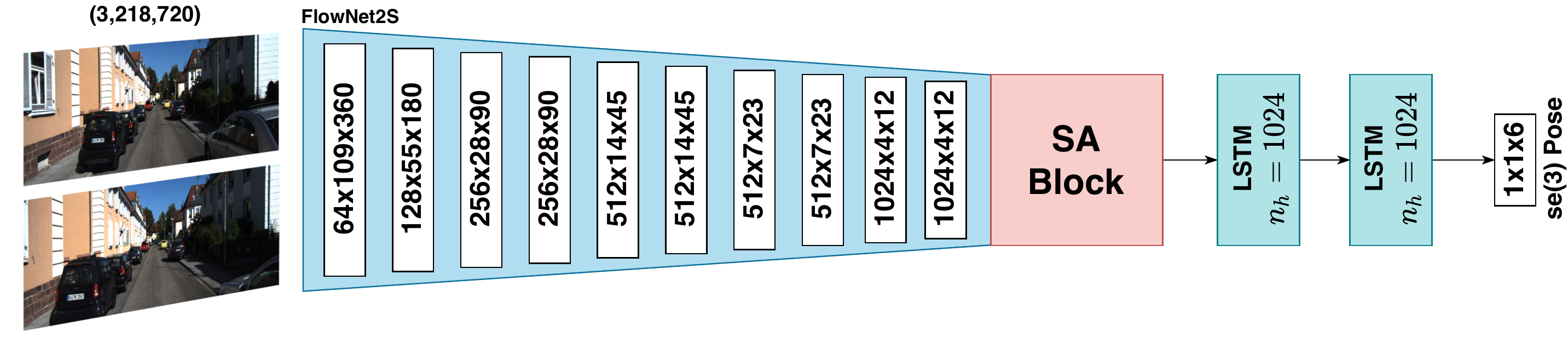}
   \caption{The overview of the proposed architecture.}
   \label{fig:overviewarch}
\end{figure*}

\section{Methodology}
\label{sec:method}
\subsection{Overview}
A general overview of the proposed architecture can be seen in Fig. \ref{fig:overviewarch}. This architecture follows a CNN-LSTM approach where the visual features corresponding to the motion between two frames are captured by a CNN and passed to a multi-layered LSTM for temporal modeling. To facilitate faster convergence, we use the pretrained weights from an optical flow estimation network~\cite{ilg2016flownet} to initialize our visual encoder. This way, the network can extend the input-output relationship formed to estimate the motion of pixels to derive the poses between the input frames. Fig. \ref{fig:overviewarch} depicts the architecture of this encoder where the input RGB images are stacked together to form a tensor of size $6 \times H \times W$ and passed through 10 stages of convolutional layers to extract the necessary visual features. 

\begin{figure}[b]
    \hspace{1.3cm}
   \includegraphics[width=0.65\linewidth]{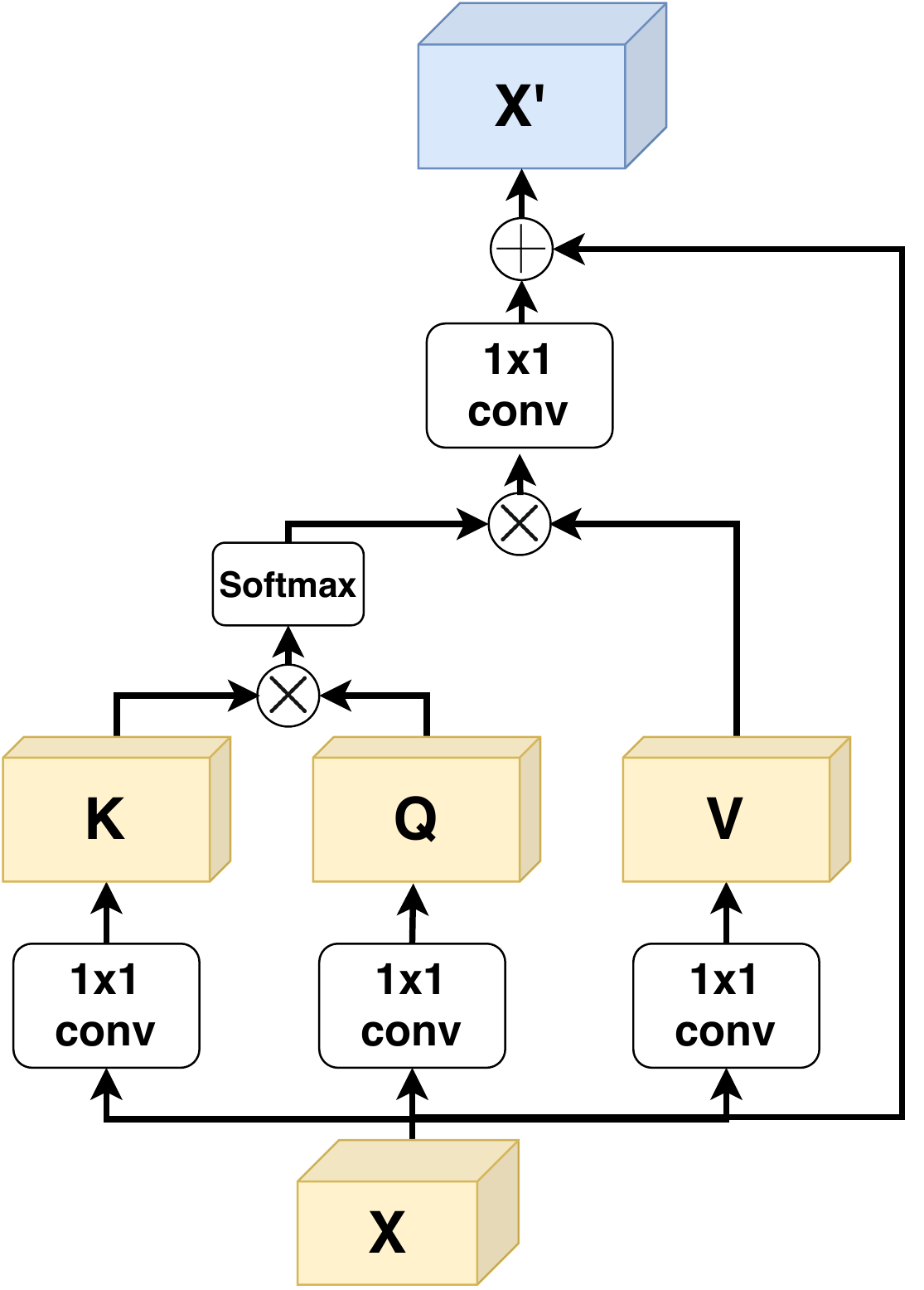}
   \caption{Detailed structure of the self-attention block.}
   \label{fig:sablock}
\end{figure}

\subsection{Self-Attention Block}
The extracted features are then passed through the self-attention block, which outputs feature maps with the same size as the inputs. Our implementation of this block is based on \cite{zhang2018sagan} and is visualized in Fig. \ref{fig:sablock}. The input to the SABlock is first linearly projected to form three feature maps, named query, key, and value, through independent weight matrices as follows:

\begin{equation}
\mathbf{Q} = \mathbf{W}_q\mathbf{X}, \mathbf{K} = \mathbf{W}_k\mathbf{X}, \mathbf{V} = \mathbf{W}_v\mathbf{X}
\end{equation}

Where $\mathbf{X} \in \mathbb{R}^{d_{in} \times n}$, $\mathbf{W}_q, \mathbf{W}_k \in \mathbb{R}^{d_k \times d_{in}}$, and $\mathbf{W}_v \in \,\mathbb{R}^{d_v \times d_{in}}$.~Then, to calculate the unnormalized attention map, the query and key matrices are multiplied and passed through a softmax function. 

\begin{equation}
\boldsymbol{\lambda} = softmax(\frac{\mathbf{Q}^T\mathbf{K}}{\sqrt{\mathbf{d_k}}}), \boldsymbol{\lambda} \in \mathbb{R}^{n \times n}
\end{equation}

The multiplication step weighs the encoded content of the input features from every region against the encoded contents of the other regions, yielding the said attention map. The purpose of the subsequent softmax function is to normalize the attentiveness of every region to each query vector. In other words, the softmax function normalizes $\boldsymbol{\lambda}$ such that the rows of the $n$ by $n$ attention map sum to one. This formulation resembles a weighted averaging process where the weights are derived based on the similarity of the content between input features. Thereafter, the resulting attention map is used alongside the value matrix to derive the weighted features. 

\begin{equation}
\mathbf{O}_{h_{i}} = \boldsymbol{\lambda} \mathbf{V}^T,
\end{equation}

Where $\mathbf{O}_{h_{i}}$ corresponds to the output of a single head of this attention mechanism. Since every feature vector derived from the visual encoder before the SABlock has a limited receptive field of the input images, an attention-less network would have no way of discriminating against features that contain erroneous information. However, by passing the said features through the self-attention block, the model can manipulate the faulty features and rectify the derived motion representation based on the global content of the input. The outputs from each of the heads are then concatenated and multiplied by a weight matrix $\mathbf{W}_o$ to yield the output of the SABlock. Thereafter, this output is weighted by a learnable parameter and the results are treated as residual values by being added to the original feature maps. This process is formulated as below

\begin{equation}
    \mathbf{O} = \overset{h}{\underset{i=1}{\mathrm{concat}}}[\mathbf{O}_{h_i}] \\
    \label{eq:1}
\end{equation}
\begin{equation}
    \mathbf{X}' = \mathbf{X} + \boldsymbol{\gamma} \times (\mathbf{W}_o\mathbf{O})
    \label{eq:2}
\end{equation}

Where h is the number of heads and $\boldsymbol{\gamma}$ is a learnable parameter that is initialized to zero. In practice we set h~to~1 throughout this work.

\subsection{Pose Estimation}
The output from the self-attention block is then passed through a global average pooling (GAP) layer to derive the vector containing the motion information,

\begin{equation}
    \mathbf{M} = \frac{1}{H\times W}\sum_{i=1,j=1}^{H, W}\mathbf{X}\textprime_{ij}^{c}
\end{equation}

Where $c$ represents the depth index of the feature map. The utilization of the SABlock before the GAP layer allows for a more refined dimensionality reduction process. This compares to a setting where SABlocks are not used before the GAP layer and both the accurate and inaccurate information are naively averaged, leading to reduced feature maps that encode warped representations of the transformation between the input images. This is a crucial point in the case of visual odometry where feature reliability is essential for accurate odometry estimation. The lack of mechanisms which provide a global sense of motion that is present in the input, leads to the inability of the network in estimating the true motion when artifacts are present in the input.

Thereafter, as can be seen in Fig.~\ref{fig:overviewarch}, the resulting vector from the GAP layer, $\mathbf{M}$, is used as input to a two-layer LSTM network to model the motion vectors in a temporal sense. A two-layer fully connected network, which is not shown in Fig. \ref{fig:sablock}, is used to derive the 6-DoF pose that represents the incremental movement between the image pairs.

\subsection{Loss Function}
In our approach, we parameterize the desired output of the network as $\boldsymbol{\xi}^*=ln(\mathbf{T}^*)$ where $\mathbf{T}^*$ represents the true $4\times 4$ transformation matrix between the global poses of the two input images. Therefore, the network is trained to estimate $\boldsymbol{\xi}^* \in \mathbb{R}^{6}$, the lie algebra coordinates, which is the logarithm mapping of $\mathbf{T}^*$. Similar to \cite{valentin2018dpcnet}, the loss function itself is formulated as follows

\begin{equation}
    \mathcal{L} = \frac{1}{2}\mathbf{g}(\boldsymbol{\xi})^T(\boldsymbol{\Sigma}^{-1})\mathbf{g}(\boldsymbol{\xi})
\end{equation}
Where
\begin{equation}
    \mathbf{g}(\boldsymbol{\xi}) = \text{ln}(\text{exp}(\boldsymbol{\xi})\mathbf{T}^{*{^{-1}}})
\end{equation}
Where $\boldsymbol{\xi}$ is the output of the network, and $\mathbf{T}^*$ is the true transformation matrix. To weigh the motion loss in each of the axes and balance their mutual impact, we use the empirical covariance matrix which is calculated using the training set through the following formulation

\begin{equation}
\mathbf{\mathbf{\Sigma}}=\frac{1}{N-1} \sum_{i=1}^{N}\left(\boldsymbol{\xi}_{i}^{*}-\overline{\boldsymbol{\xi}^{*}}\right)\left(\boldsymbol{\xi}_{i}^{*}-\overline{\boldsymbol{\xi}^{*}}\right)^{\mathbf{T}}
\end{equation}

where $\overline{\boldsymbol{\xi}^{*}} = \frac{1}{N} \sum_{i=1}^{N} \boldsymbol{\xi}_{i}^{*}$ and $\boldsymbol{\xi}^* = ln(\mathbf{T}^*)$.

\subsection{Architectural Details}
As can be seen in Fig. \ref{fig:overviewarch}, the CNN section of our architecture is composed of 10 layers. The three initial layers of our CNN use kernels with sizes of $7\times7$, $5\times5$ and $5\times5$, respectively, while each uses a stride of 2. All the preceding layers have strides 1 and 2 intermittently. The number of output channels for each layer is depicted in Fig. \ref{fig:overviewarch}. For the SABlock, we set $d_k$, $d_q$ and $d_v$ to 128. Both of the LSTM layers are made up of 1024 hidden units while a single layer fully connected layer is used to estimate the output.

\section{Implementation and Evaluation}
\label{implementation}
We used PyTorch and PyTorch lightning to implement our method. An NVIDIA Tesla T4 GPU was used to train all the networks. The networks were trained for 300 epochs, while early stopping techniques were employed to reduce overfitting. Furthermore, a learning rate of 1e-4 and Adam optimizer \cite{kingma2015AdamAM} with the proposed hyperparameters in \cite{kingma2015AdamAM} were used for training.

\subsection{Datasets}
For all experiments, we used the KITTI odometry dataset. This dataset consists of 22 sequences of driving a car in urban areas in the presence of a large number of moving objects and artifacts in the images. The ground truth for the first 11 of the sequences is available and we use sequences 0, 1, 2, 8, and 9 for training and leave 3, 4, 5, 6, 7, and 10 for testing. Random short sequences with random lengths are chosen from the training sequences while 5\% of them are separated for validation purposes. The images are resized to (218, 720) and the training sequences are shuffled at the end of each epoch. It should be noted that no augmentations were done on the input images.

\subsection{Evaluation Metrics}
In order to provide quantitative measures and compare the performance of our approach against the classical and data-driven methods, we use the KITTI odometry benchmarking methods proposed in \cite{geiger2012kitti}. This evaluation method reports the relative translation and rotation accuracy of the estimates over sequences of length 100m-800m. The translation accuracy is reported as the percentage of deviation from the ground truth, while rotation error is reported in units of degrees per-meter. Averaging the error over different distances provides an adequate measure of drift as well as incremental accuracy.

\section{Experiments}
\label{experiments}
\subsection{Quantitative Results}
The quantitative results of our approach are reported in Table \ref{tab:results}. The competing SOTA approach in deep learning based visual odometry is DeepVO \cite{wang2017deepvo} while VISO2-M \cite{geiger2011viso}, a SOTA feature-based algorithm, is chosen as the classical approach that we will compare our results against. Since the codes for DeepVO were not open-sourced, we implemented this method as close as possible based on \cite{wang2017deepvo}. It should be noted that the structure of DeepVO closely resembles our network with the exception of the attention layers. As can be seen in Table \ref{tab:results}, our average and sequence-based results consistently surpass DeepVO. In particular, the averaged accuracy of our method surpass that of DeepVO by 32.4\% on translation and 37.7\% on rotation, proving the advantage of attention on VO in terms of performance. Additionally, the performance of our method is consistently better than the monocular vision based VISO2 in terms of translation accuracy with the exception of sequence 6. In terms of rotation, our method outperforms VISO2-M in sequences 5, 7 and 10. Overall, the average results on all test sequences show that our method is able to surpass the performance of both classical and SOTA deep learning based odometry baselines.

\begin{table}[t]
\caption{Quantitiative results against competing approaches}
\vspace{0.3cm}
\label{tab:results}
\centering
\resizebox{\columnwidth}{!}{%
\begin{tabular}{c || c || c || c}
  \toprule
  \multirow{2}*{Test Seq.}   & VISO2-M \cite{geiger2011viso} & DeepVO \cite{wang2017deepvo} & \textbf{AVO (ours)}\\
       & t\textsubscript{rel}(\%)/r\textsubscript{rel}($^{\circ}$)& t\textsubscript{rel}(\%)/r\textsubscript{rel}($^{\circ}$)& t\textsubscript{rel}(\%)/r\textsubscript{rel}($^{\circ}$) \\
  \hline 
  Seq 03    & $27.63/\mathbf{0.055}$  & $12.88/0.056$  & $\mathbf{12.29}/0.073$  \\
  \hline 
  Seq 04    & $15.82/\mathbf{0.018}$  & $13.26/0.048$  & $\mathbf{11.31}/\mathbf{0.018}$    \\
  \hline 
  Seq 05    & $16.54/0.055$  & $11.49/0.037$  & $\mathbf{7.662/0.028}$  \\
  \hline 
  Seq 06    & $\mathbf{10.14}/\mathbf{0.027}$  & $17.73/0.060$  & $12.17/0.037$            \\
  \hline 
  Seq 07    & $33.23/0.154$  & $13.47/0.090$  & $\mathbf{7.582/0.034}$           \\ 
  \hline 
  Seq 10    & $29.18/0.115$  & $18.18/0.073$  & $\mathbf{14.66}/\mathbf{0.040}$           \\ 
  \bottomrule
  Avg.      & $22.09/0.071$  & $14.50/0.061$  & $\mathbf{10.95}/\mathbf{0.038}$           \\ 
  \bottomrule
\end{tabular}
}
\end{table}

\subsection{Qualitative Results}
\begin{figure*}
\hfill
\subfigure[Sequence 3]{\includegraphics[width=4.28cm]{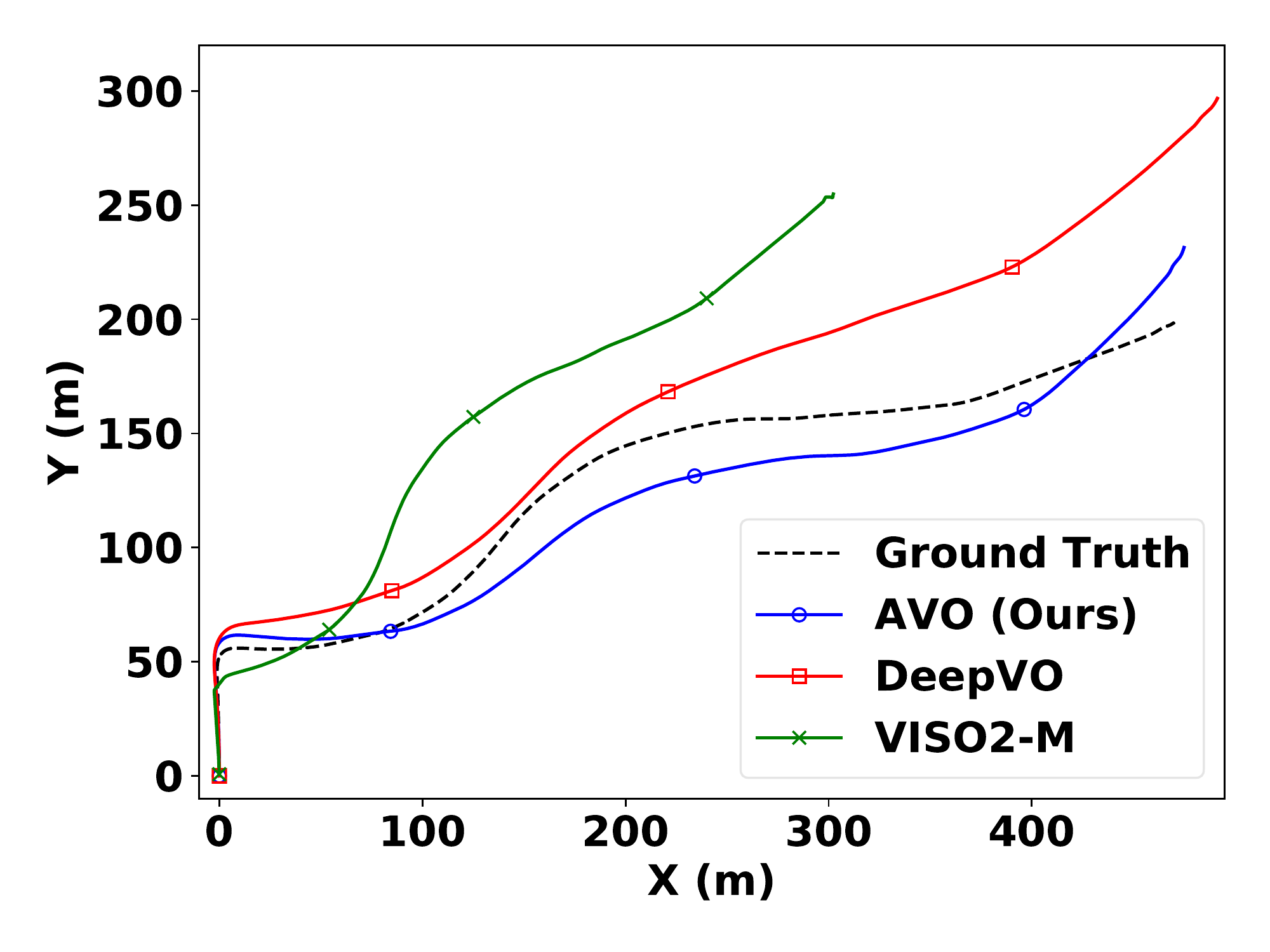}\label{fig:qual3}}
\hfill
\subfigure[Sequence 5]{\includegraphics[width=4.28cm]{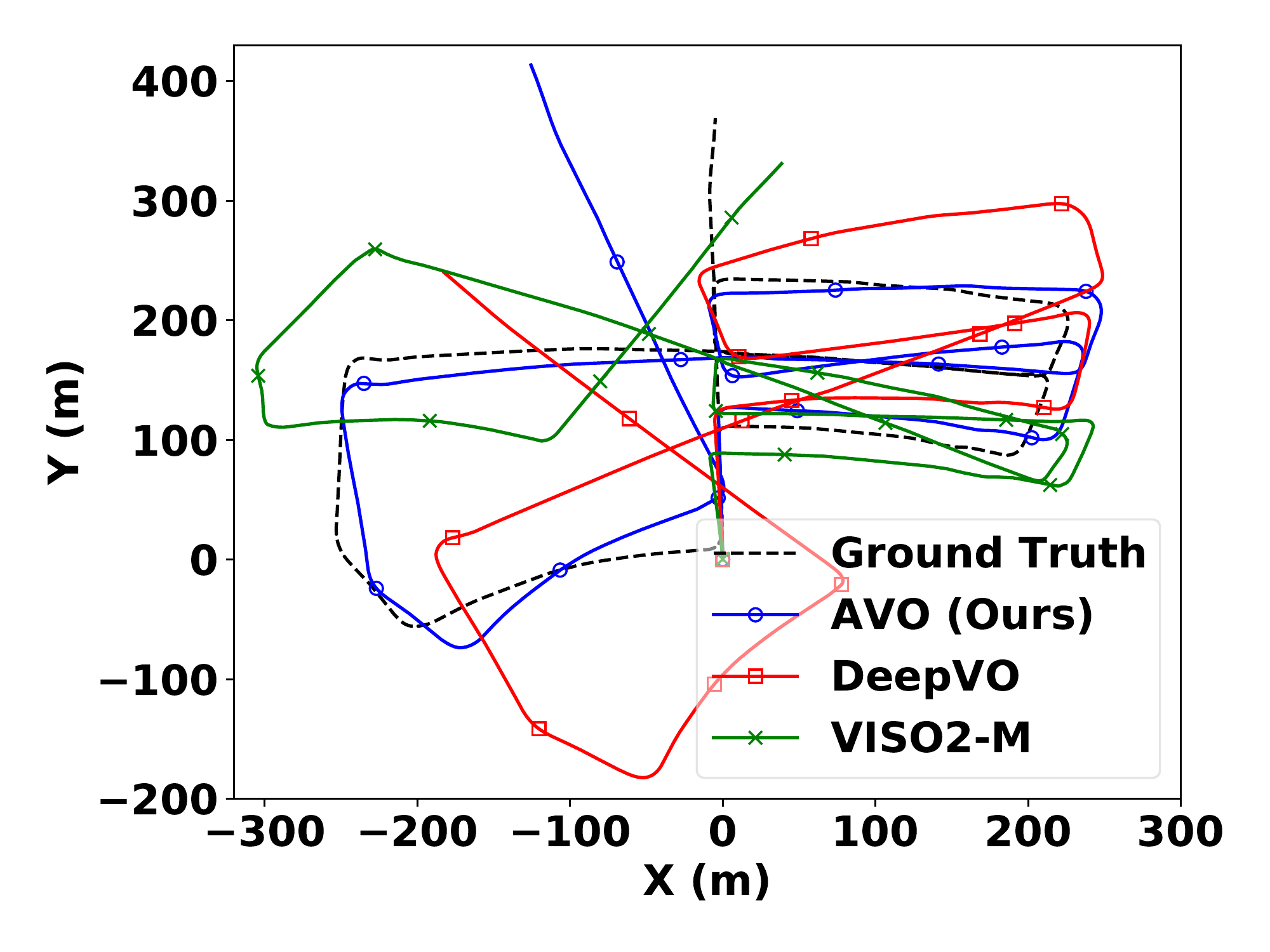}\label{fig:qual5}}
\hfill
\subfigure[Sequence 7]{\includegraphics[width=4.28cm]{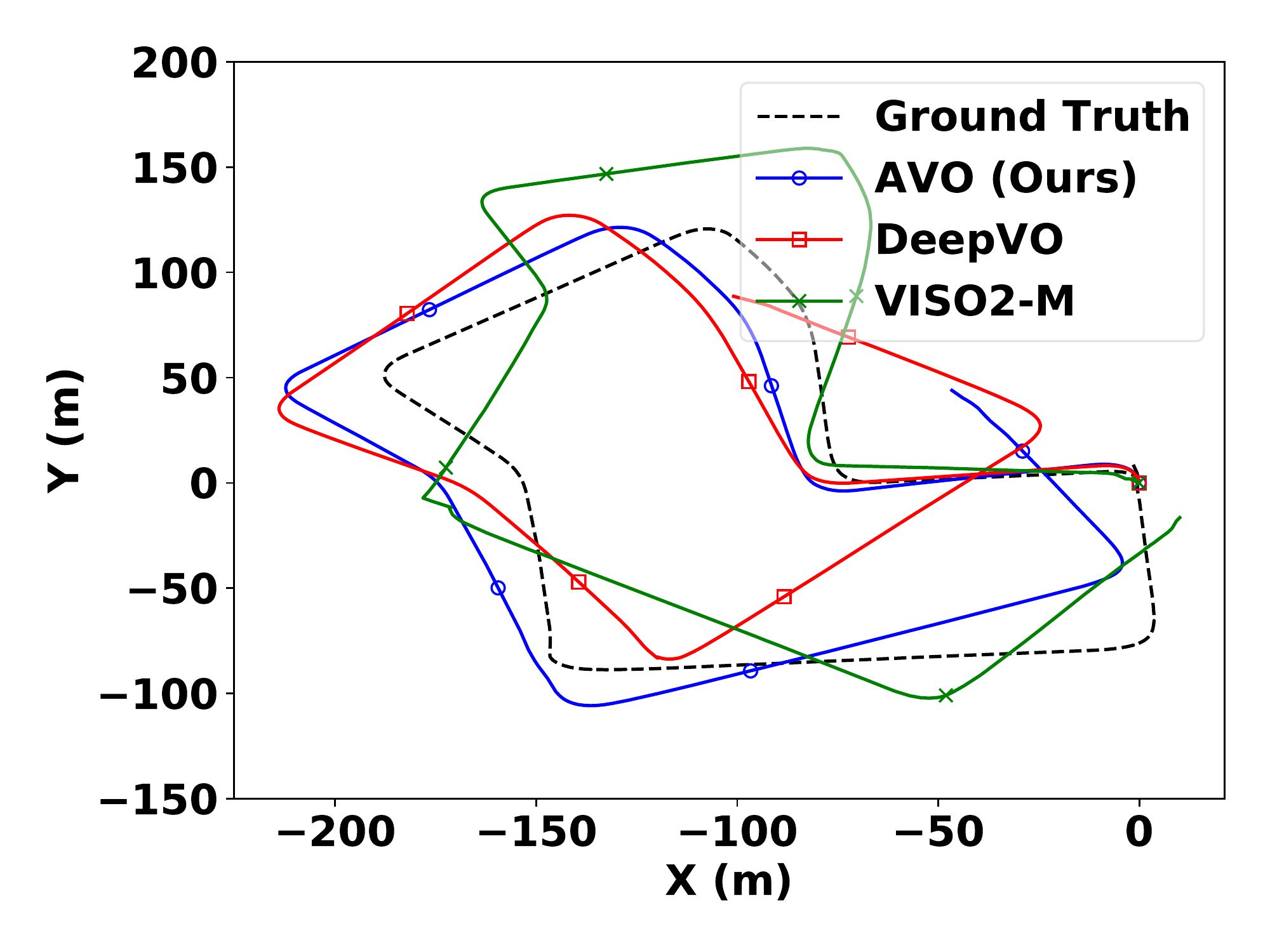}\label{fig:qual7}}
\hfill
\subfigure[Sequence 10]{\includegraphics[width=4.28cm]{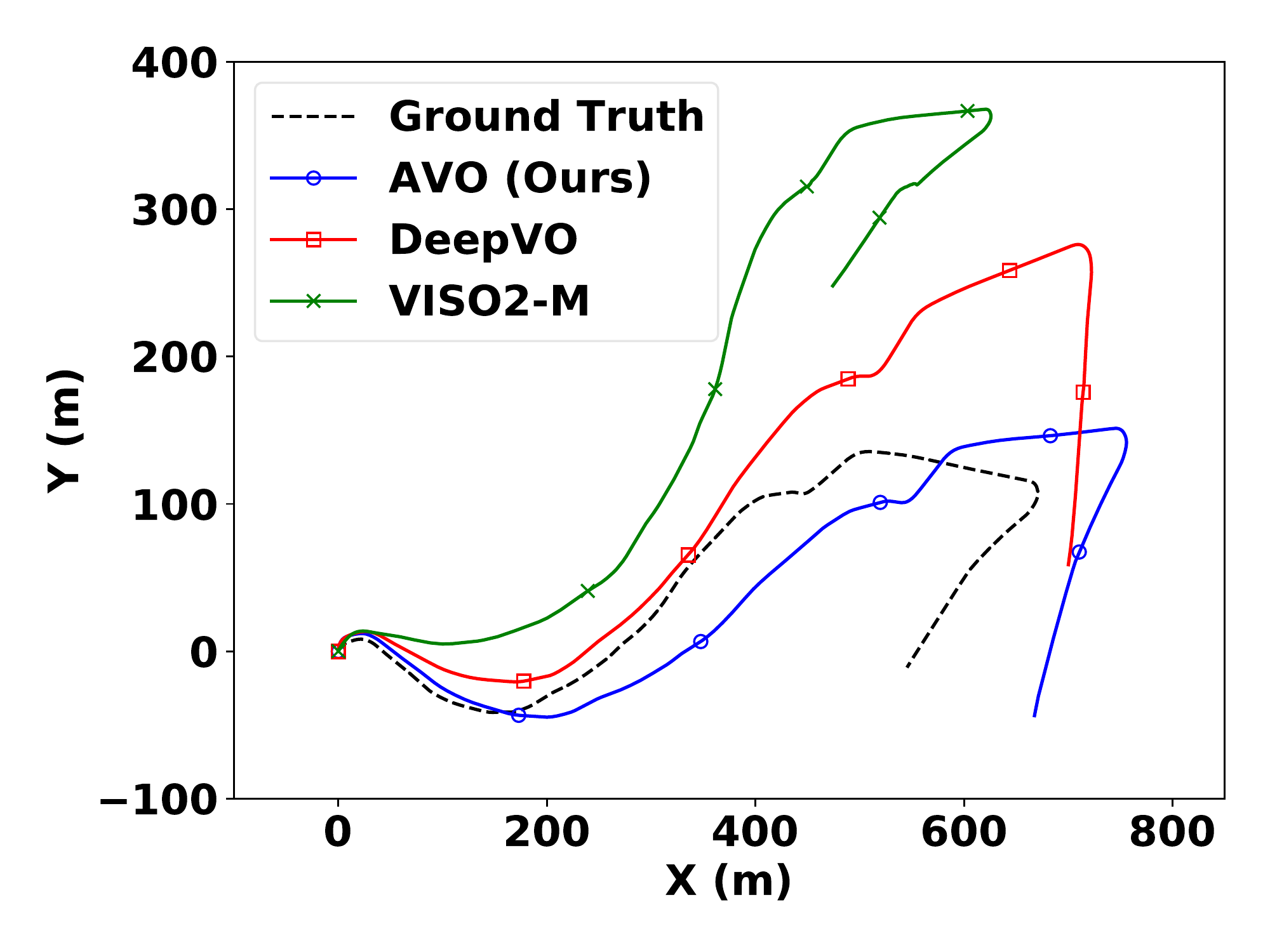}\label{fig:qual10}}
\hfill
\caption{Qualitative results for the test sequences of the KITTI dataset.}
\label{fig:qualitative}
\end{figure*}

The visualizations regarding the trajectory estimates are provided in Fig. \ref{fig:qualitative}. It can be seen that our network is able to track the ground truth trajectory significantly better in case of sequence 5. This sequence alongside sequence 7 contain numerous moving objects present in the scene while sequence 5 also exhibits tortuous paths. Furthermore, based on the results from sequence 10 and 3, the issue of drift in DeepVO~\cite{wang2017deepvo} is apparent from the initial segment of the trajectories while our approach drifts slower. Compared to VISO2-M~\cite{geiger2011viso} and based on results from Fig.~\ref{fig:qual3} and Fig.~\ref{fig:qual10}, our network is able to maintain a slower drift while providing a better accuracy in terms of tracking the true trajectory. Based on Fig. \ref{fig:qual3} and Fig. \ref{fig:qual10}, VISO2-M, is not able to estimate the scale properly while our method shows an adequate implicit scale estimation capability that is formed during training. Additionally, VISO2-M shows an unstable behavior, during a short stopping of the car, based on the estimated path at location~(-175,-10) of Fig. \ref{fig:qual7} while none of the deep learning based methods exhibit such behavior. 

\subsection{Attribution Analysis}
\label{sec:attrs}
To provide an insight into how self-attention allows the network to focus on certain regions of the image and to show the ability of the attention mechanism in manipulating the way features are utilized, we use integrated gradients \cite{sundararajan2017ig} to provide a saliency analysis of our approach. Through this method, we visualize the gradient of the output pose with respect to each of the pixels of the input images. In other words, we will be able to qualitatively analyze the effect of each region of the input on the output pose. Therefore, by comparing the generated saliency maps from our approach against DeepVO~\cite{wang2017deepvo}, we will be able to get an intuition into the effects of attention modules on the feature extraction capabilities of VO networks.

\begin{figure*}[hbt!]
\hfill
\subfigure[Saturation]{\includegraphics[width=5.5cm]{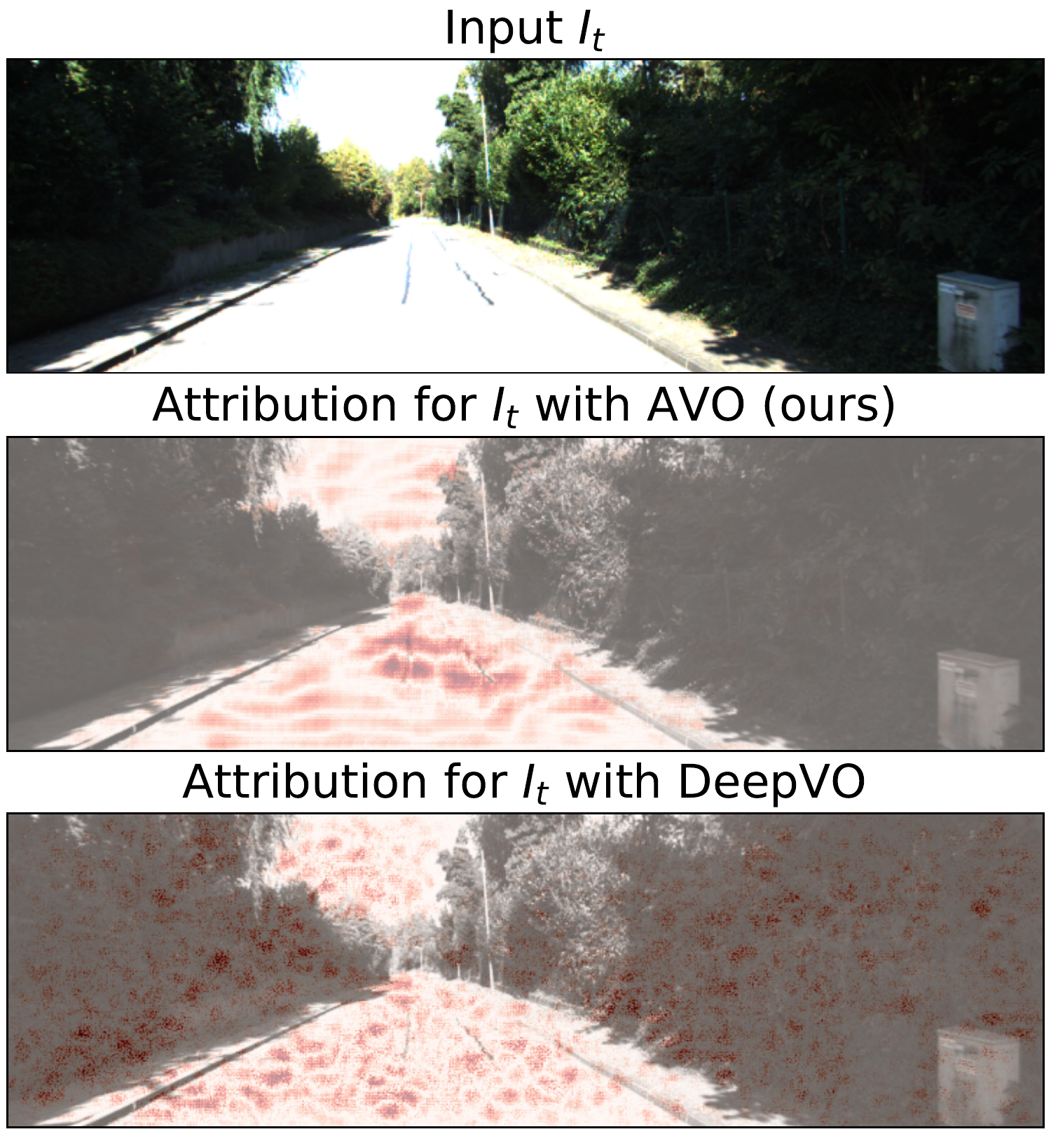}\label{fig:attrsaturation}}
\hfill
\subfigure[Turning]{\includegraphics[width=5.5cm]{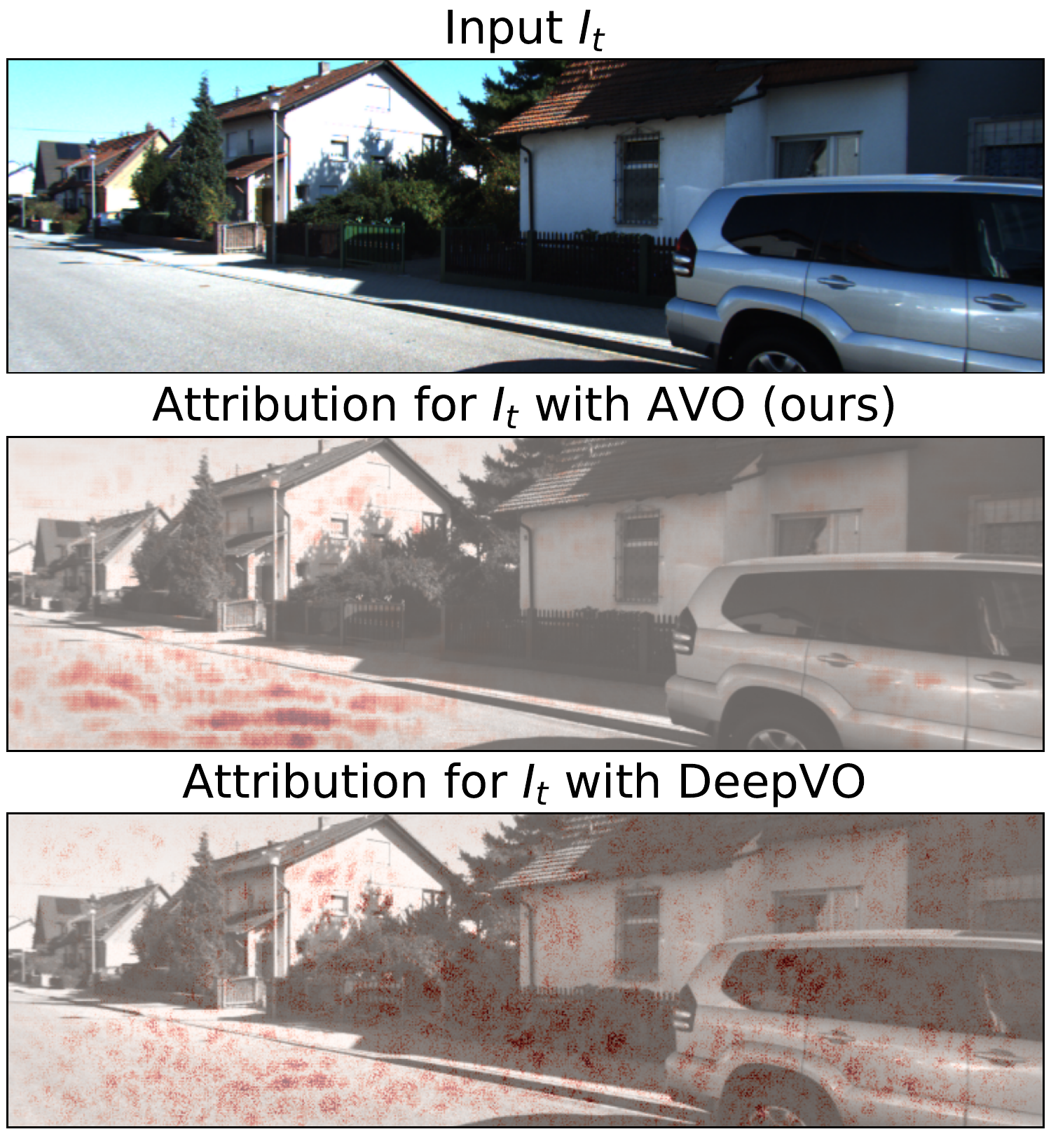}\label{fig:attrturning}}
\hfill
\subfigure[Straight Movement]{\includegraphics[width=5.5cm]{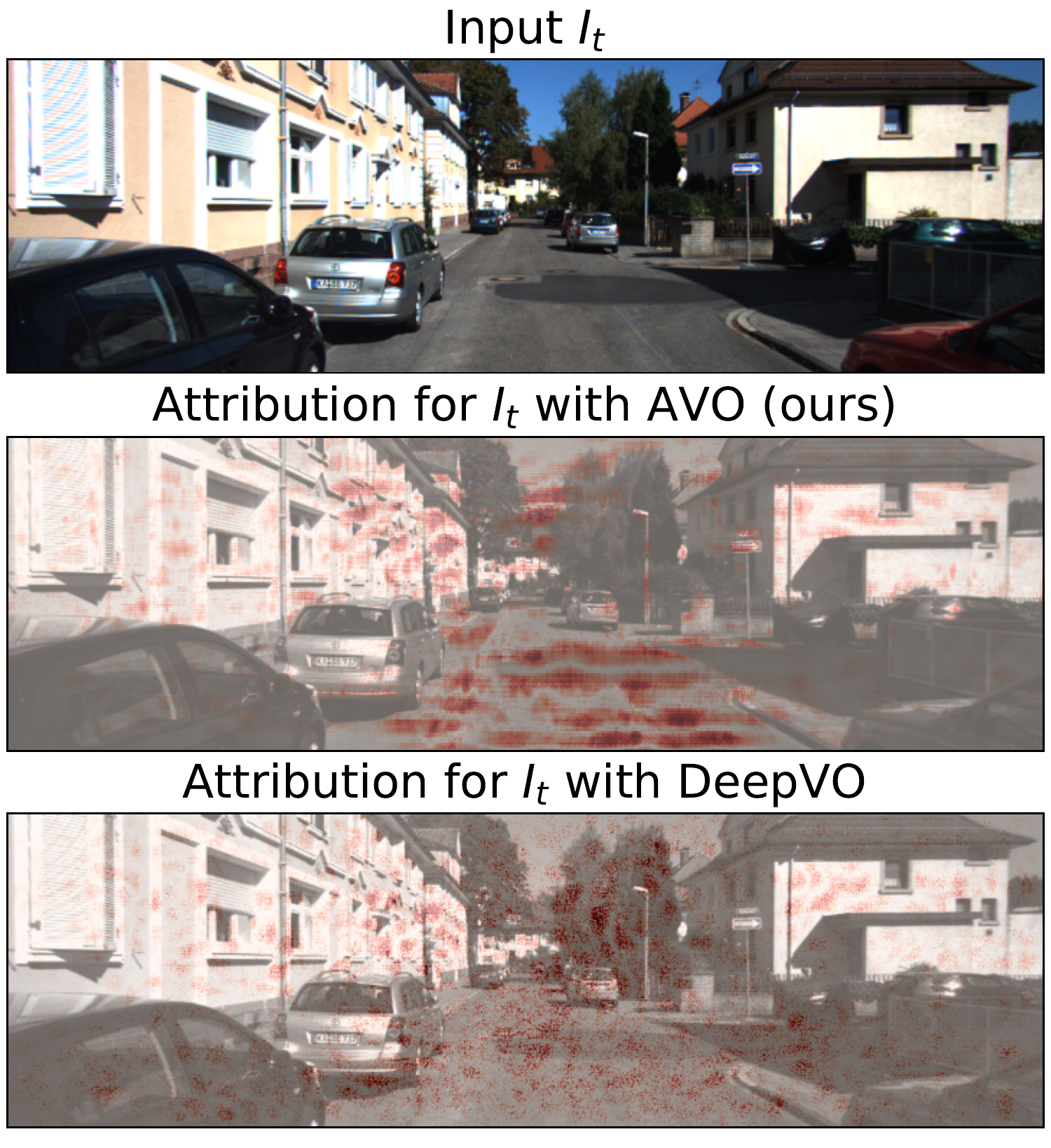}\label{fig:attrstraight}}
\hfill
\vfill
\hfill
\subfigure[Moving Object 1]{\includegraphics[width=5.5cm]{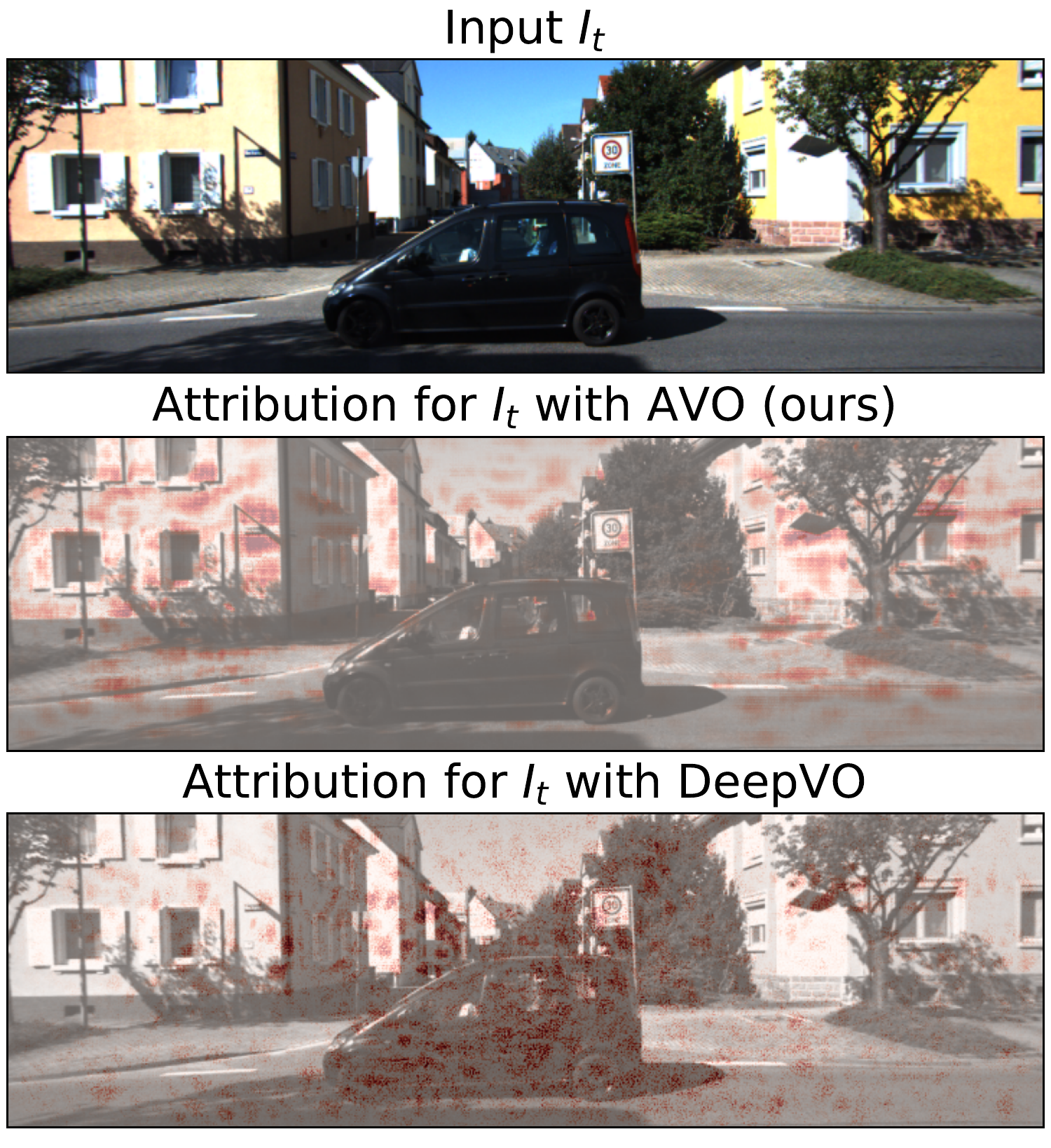}\label{fig:attrmoving1}}
\hfill
\subfigure[Moving Object 2]{\includegraphics[width=5.5cm]{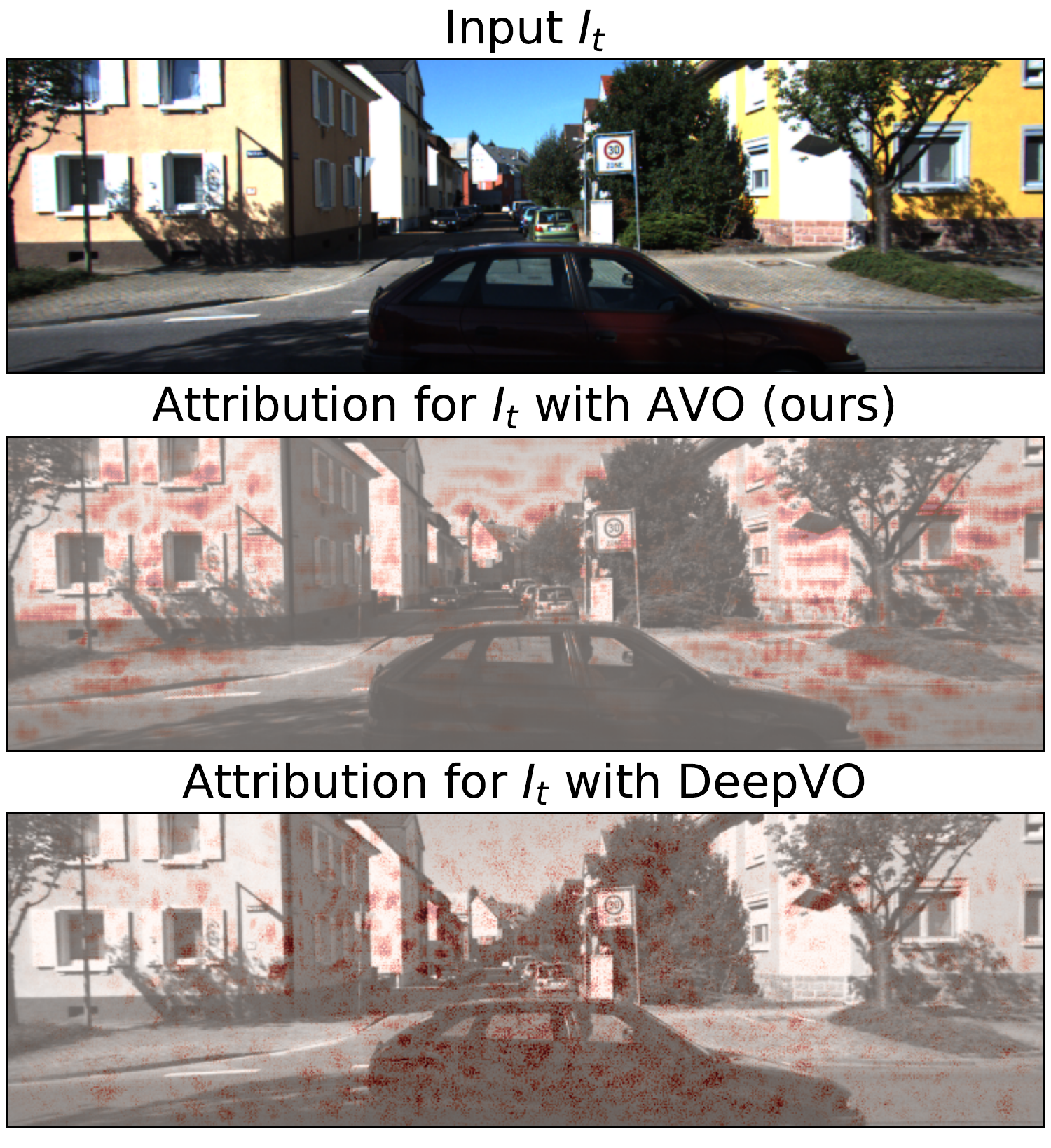}\label{fig:attrmoving2}}
\hfill
\subfigure[Moving Object 3]{\includegraphics[width=5.5cm]{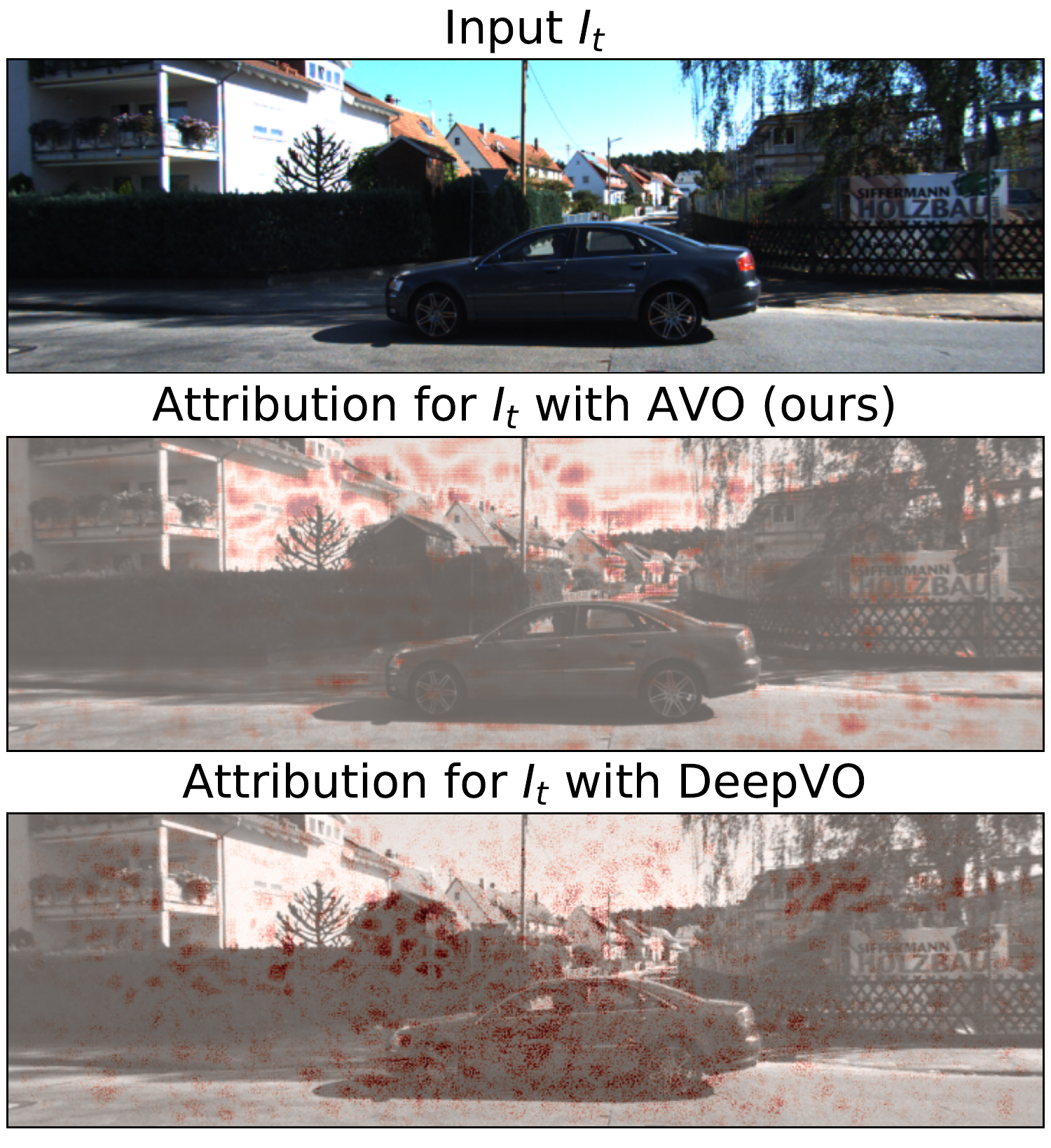}\label{fig:attrmoving3}}
\hfill
\caption{Attribution analyses using integrated gradients~\cite{sundararajan2017ig} in different scenarios (best viewed in color)}
\label{fig:attrs}
\end{figure*}

The results of the attribution analyses are presented in Fig. \ref{fig:attrs}. As can be seen in this figure, 6 scenarios are presented where three contain moving objects in the scene, and the others correspond to straight movement, turning, and saturated images. The intensity of the red pixels in the overlaid figures on the bottom rows of Fig. \ref{fig:attrs} show the gradient magnitudes of those pixels with respect to the output pose. Based on Fig. \ref{fig:attrmoving1}, Fig. \ref{fig:attrmoving2}, and Fig. \ref{fig:attrmoving3} it can be seen that our approach is able to consistently disregard the information from moving objects and use the salient features from the background to make estimations regarding the changes in the pose of the vehicle. This is while DeepVO seems to have no structure in terms of dynamically changing the effect of different pixels on the output. The same patterns present themselves when analyzing Fig. \ref{fig:attrstraight} and Fig. \ref{fig:attrturning} where our network focuses on foreground objects consistently across different scenarios while disregarding artifacts in the input. In particular, our network puts a strong emphasis on the usage of pixels that belong to the road and buildings to provide pose estimates even in the presence of saturation, as can be seen in Fig. \ref{fig:attrsaturation}.

\subsection{Loss Analysis}
\begin{figure}[h]
\hfill
\subfigure[Sequence 3]{\includegraphics[width=8.cm]{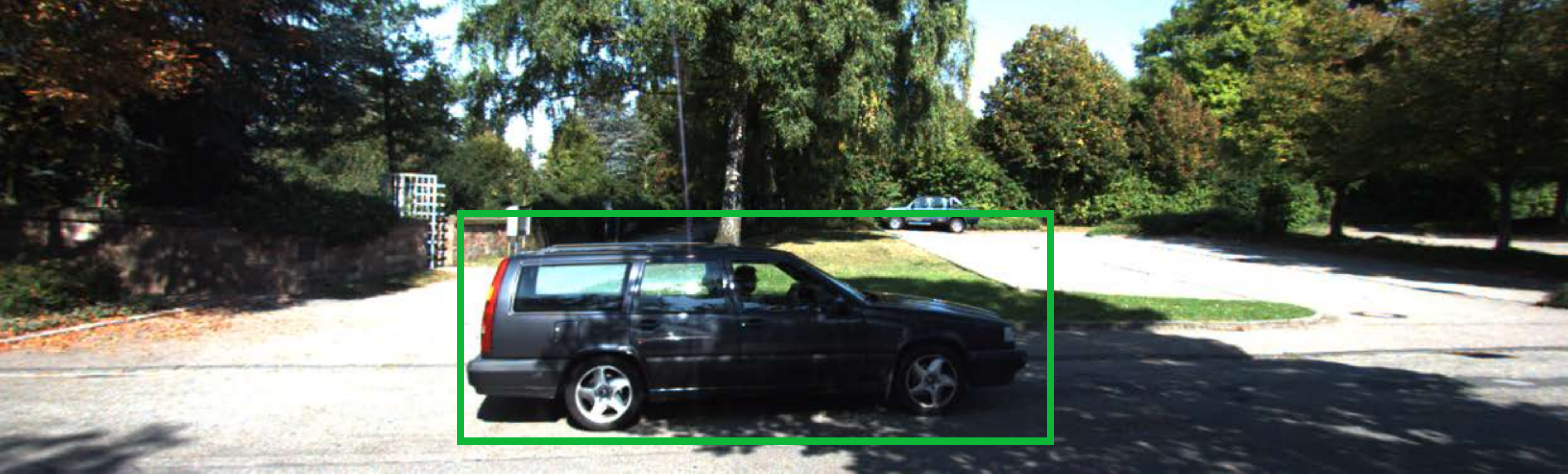}\label{fig:lossconsistent}}
\hfill
\vfill
\hfill
\subfigure[Sequence 7]{\includegraphics[width=8.cm]{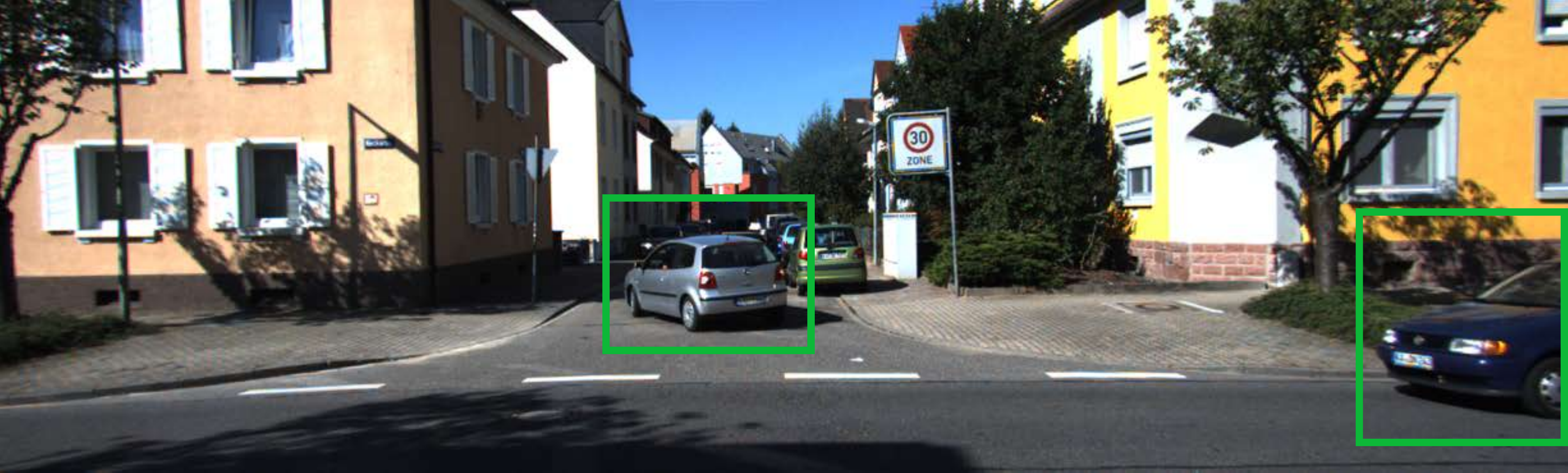}\label{fig:lossentrance}}
\hfill
\vfill
\hfill
\subfigure[Loss plot for Seq. 3]{\includegraphics[width=4.cm]{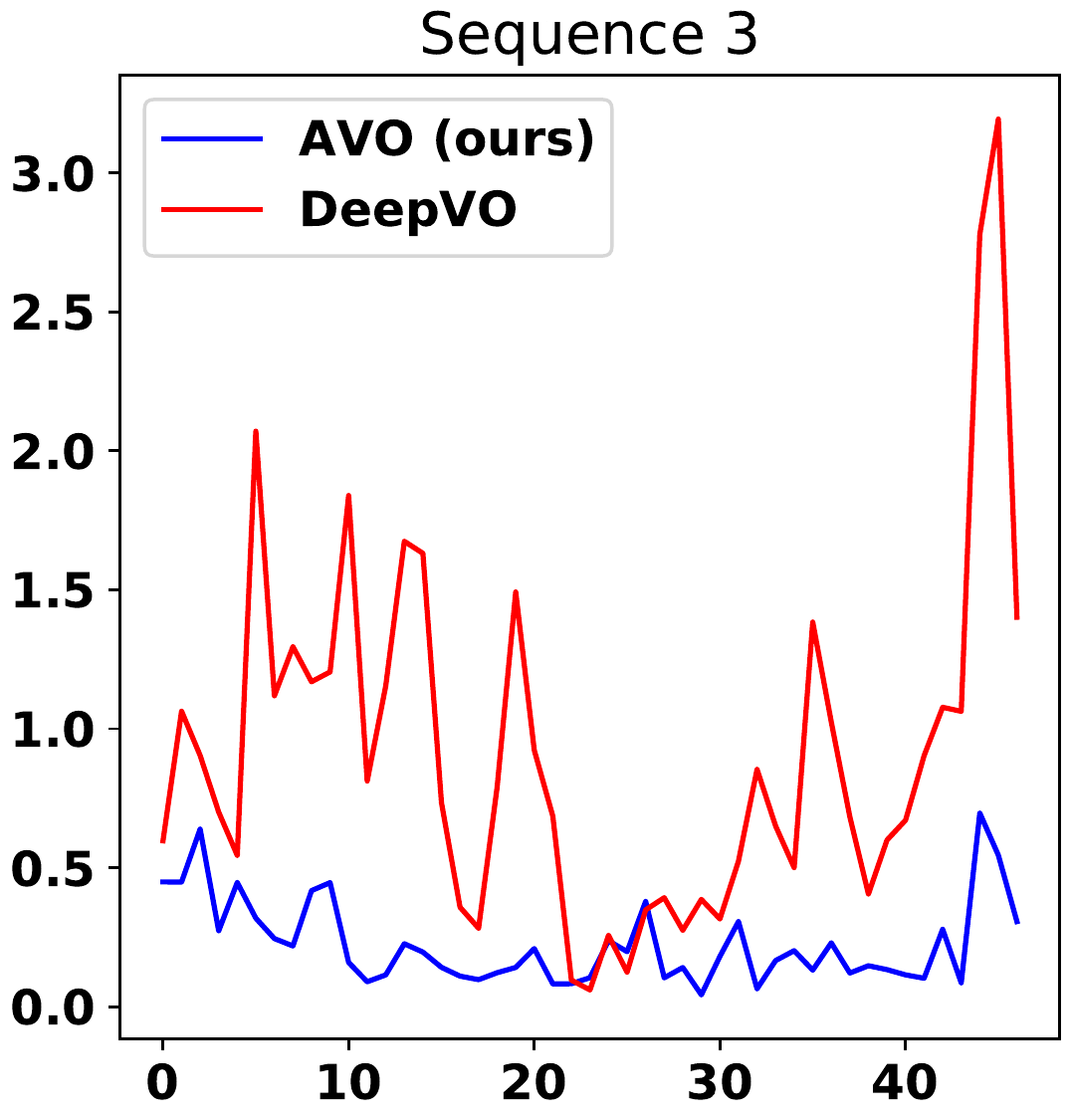}\label{fig:lossconsistentplot}}
\hfill
\subfigure[Loss plot for Seq. 7]{\includegraphics[width=4.cm]{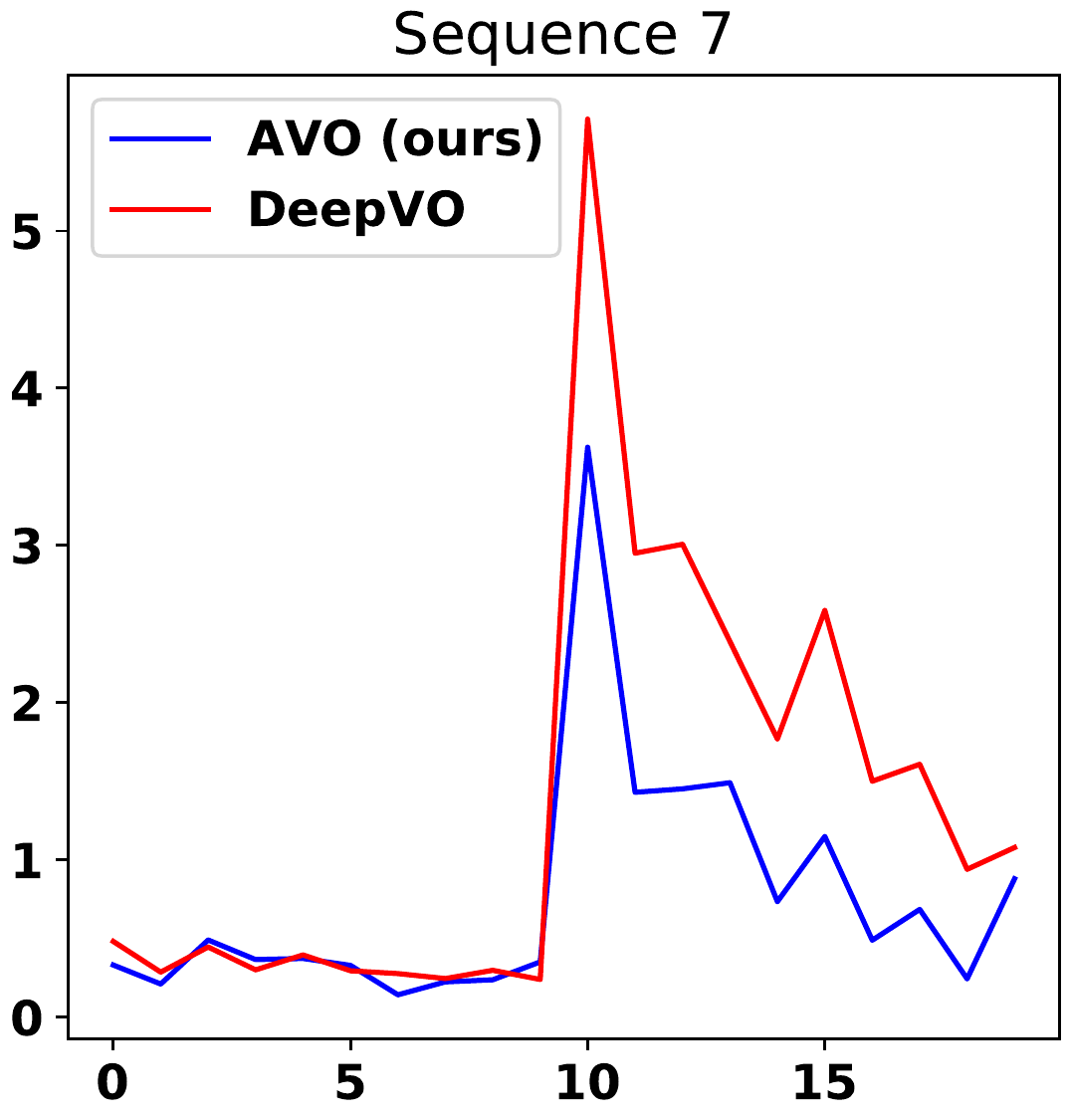}\label{fig:lossentranceplot}}
\hfill
\caption{Input pairs alongside frame-to-frame loss plots for test sequences (best viewed in color)}
\label{fig:loss}
\end{figure}

To analyze the effect of moving cars on the accuracy of the estimates from each of the networks, we visualized the frame-to-frame loss values in the presence of moving cars in the scene. Fig. \ref{fig:lossconsistent} shows a moving object that was present in the scene in test sequence 3 of the KITTI dataset. We visualize the loss from each of the networks during the presence of this object in the input frames. As can be seen in Fig. \ref{fig:lossconsistentplot}, our network is able to maintain its accuracy throughout this scene while exhibiting minimal amounts of changes in the value of error. This is while the loss from DeepVO~\cite{wang2017deepvo} fluctuates significantly for the duration of this sequence. Combined with the results from Section \ref{sec:attrs}, it is evident that the ability of our method in rejecting moving instances allows for a smaller error as well as fewer fluctuations. Fig. \ref{fig:lossentrance} presents a case where a moving vehicle enters the frame. Based on the plots in Fig. \ref{fig:lossentranceplot}, upon the entrance of the vehicle, the loss values for both networks experience a sudden increase at index 10 of the x-axis. However, our method shows a lower value of error at the peak of this change, which eventually leads to a better performance on this test sequence.

\subsection{Consistency Analysis}
\begin{figure}[h]
\hfill
\subfigure[Top Salient categories for AVO (ours)]{\includegraphics[width=8.25cm]{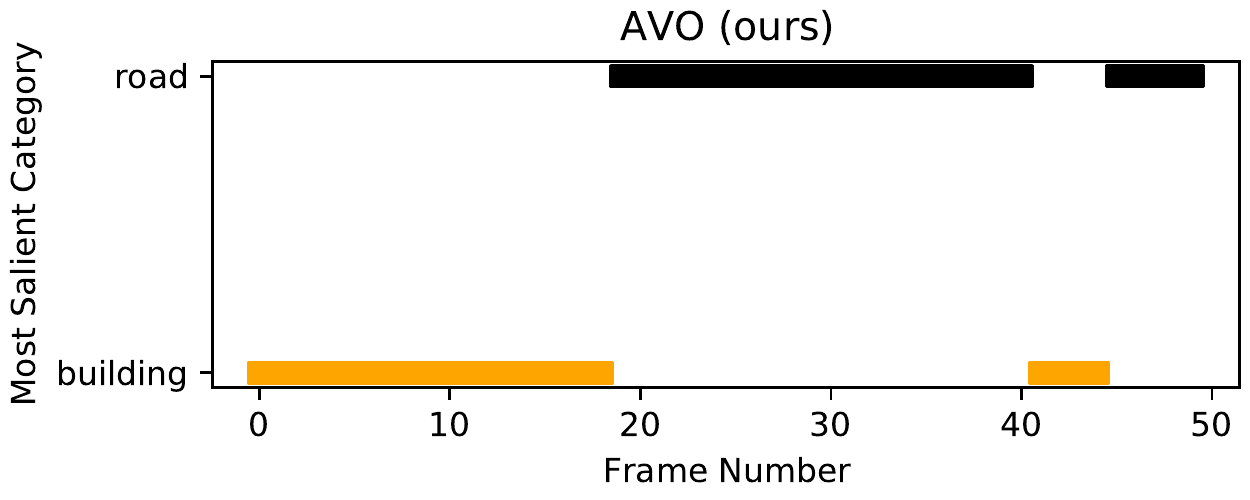}}
\hfill
\vfill
\hfill
\subfigure[Top salient categories for DeepVO~\cite{wang2017deepvo}]{\includegraphics[width=8.25cm]{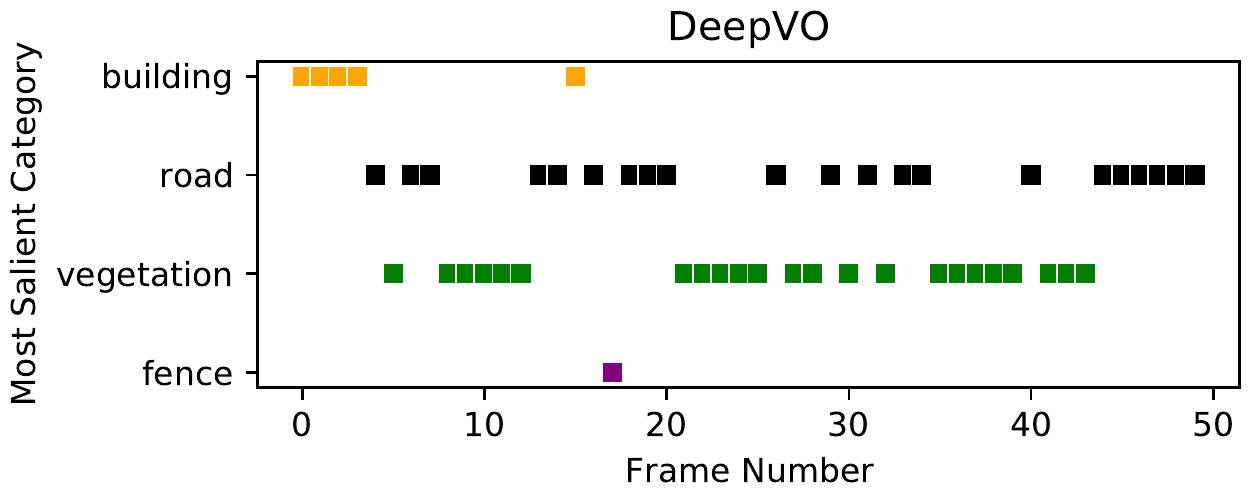}}
\hfill
\caption{Top salient categories for both our method and DeepVO through 5.}
\label{fig:consistency}
\end{figure}

In this analysis, we show the consistency of our method in extracting salient features from the inputs. To provide meaningful visualizations regarding the reliance of each model on various objects that are present in the scene, through time, we used a semantic segmentation network to label each of the pixels of the input images. Specifically, we used SSeg~\cite{Zhu2019ImprovingSS}, which currently is the SOTA approach in the task of semantic segmentation on the KITTI segmentation benchmark~\cite{Alhaija2018kittisemantic}. For each frame, the category of objects that affect the output pose the most is chosen and visualized as the top salient category. By running this pipe on 50 consecutive frames on sequence 5 of the KITTI dataset, the visualization in Fig. \ref{fig:consistency} was derived. Based on this figure, our method is able to consistently focus on the road and building regions of the inputs in order to estimate the output poses while DeepVO~\cite{wang2017deepvo} shows an unstable behavior by switching between different categories of objects. These results are aligned with visualizations in Fig. \ref{fig:attrs} where our network is able to dynamically shift focus onto background objects while DeepVO showed randomness in selection of object categories to infer the incremental pose.

\subsection{Conclusion}
In this paper, we showed the efficacy of self-attention modules on VO networks by comparing attention-based VO networks against SOTA attention-less counterparts. The quantitative analyses on the KITTI odometry dataset show that the addition of self-attention module to the visual feature extractor boosts the performance of the odometry network and allows us to surpass the performance of the SOTA attention-less network. Our saliency analyses showed that with the addition of the attention module the visual feature extractor as a whole is able to dynamically reject the moving objects that are present in the scene while showing an adequate feature extraction capability even in the presence of saturation in the input images. Additionally, frame to frame loss plots showed that in the presence of moving objects, our approach is able to maintain its accuracy while SOTA methods showed unstable behaviours. Finally, semantic segmentation based analyses allowed us to further analyze the consistency of the networks and show the ability of an attention-based VO model in consistent extraction of salient features.

{\small
\bibliographystyle{ieee_fullname}
\bibliography{AVO}
}

\typeout{get arXiv to do 4 passes: Label(s) may have changed. Rerun}
\end{document}